\documentclass{llncs}

\usepackage[utf8]{inputenc}

\usepackage{graphicx} 
\usepackage[caption=false]{subfig}

\usepackage{cite}
\usepackage{nicefrac}
\usepackage{amsfonts}
\usepackage{amsmath}
\usepackage{mathtools}
\usepackage{algorithm}
\usepackage[noend]{algpseudocode}
\usepackage{wrapfig}

\newcommand{\N}{\mathbb{N}}
\newcommand{\R}{\mathbb{R}}

\newcommand{\cN}{\mathcal{N}}

\usepackage{multirow}
\usepackage[table]{xcolor}
\newcommand{\blue}{\cellcolor{blue!15}}

\usepackage{tikz}
\newcommand\copyrighttext{%
  \footnotesize  This paper is a pre-print of a paper that has been accepted for publication in the Proceedings of the 20th Pacific Asia Conference on Knowledge Discovery and Data Mining (PAKDD) 2016. 
  The final publication is available at link.springer.com   
  
  (\url{http://link.springer.com/chapter/10.1007/978-3-319-31750-2\_24}). }
\newcommand\copyrightnotice{%
\begin{tikzpicture}[remember picture,overlay]
\node[anchor=south,yshift=10pt] at (current page.south) {\fbox{\parbox{\dimexpr\textwidth-\fboxsep-\fboxrule\relax}{\copyrighttext}}};
\end{tikzpicture}%
}

\begin{document}

\mainmatter              

\title{Adaptive Seeding for Gaussian Mixture Models}

\author{Johannes Bl\"omer \and Kathrin Bujna}


\institute{Paderborn University, 33098 Paderborn, Germany\\
\email{\{bloemer,kathrin.bujna\}@uni-paderborn.de}}

\maketitle              
\copyrightnotice

\begin{abstract}
We present new initialization methods for the expectation-maximization algorithm for multivariate Gaussian mixture models.
Our methods are adaptions of the well-known $K$-means++ initialization and the Gonzalez algorithm.
Thereby we aim to close the gap between simple random, e.g. uniform, and complex methods, that crucially depend on the right choice of hyperparameters.
Our extensive experiments indicate the usefulness of our methods compared to common techniques and methods, which e.g. apply the original $K$-means++ and Gonzalez directly, with respect to artificial as well as real-world data sets.
\end{abstract}
\section{Introduction}
Gaussian mixture modelling is an important task, e.g., in the field of cluster analysis. 
A common approach is the method of maximum likelihood for which the Expectation-Maximization (EM) algorithm \cite{Dempster77} can be applied.
The EM algorithm iteratively tries to improve a given initial mixture model and  
converges to a stationary point of the likelihood function.
Unfortunately, the likelihood function is generally non-convex, possessing many stationary points \cite{Mclachlan08}. 
The initial model determines to which of these points the EM algorithm converges \cite{Celeux15}. 

\subsection{Maximum Likelihood Estimation for Gaussian Mixtures}\label{sec:intro:mle}
A {Gaussian mixture model} ($K$-GMM)  over $\R^D$ can be described by a parameter $\theta=\{(w_k,\mu_k,\Sigma_k)\}_{k=1,\ldots,K}$, where $w_k\in\R$ is the mixing weight ($\sum_{k=1}^K w_k=1$), $\mu_k\in\R^D$ is the mean, and $\Sigma_k\in\R^{D\times D}$ is the covariance matrix of the $k$-th mixture component. 
Its probability density function is given by 
$ \cN(x|\theta) = \sum_{k=1}^K  w_k \cN(x|\mu_k,\Sigma_k)$, 
where we denote the $D$-variate Gaussian distribution by $\cN(\cdot|\mu,\Sigma)$. 
Given a data set $X\subset\R^D$, the {Maximum Likelihood Estimation} (MLE) problem is to find a $K$-GMM $\theta$ that maximizes the likelihood 
$\mathcal{L}(\theta|X) = \prod_{x\in X} \cN(x|\theta)$.
For $K=1$, there is a closed-form solution \cite{Bishop06}.  
For $K>1$, the EM algorithm, whose outcome heavily depends on the initial model,  can be applied.

\subsection{Related Work}\label{sec:intro:relatedwork}
A common way to initialize the EM algorithm is to first draw means uniformly at random from the input set and then to approximate covariances and weights \cite{Biernacki03,Melnykov11,Maitra09,Meila98}.
To compensate for the random choice of initial means, several candidate solutions are created and the one with the largest likelihood is chosen. 
Often, few steps of the EM, Classification EM, or Stochastic EM algorithm are applied to the candidates. 
Similarly, the $K$-means algorithm may be used \cite[p. 427]{Bishop06}. 
Due to the random choice of the initial means, all these methods are better suited for spherical and well-separated clusters.
Furthermore, testing several candidates is computationally expensive.

Other popular initializations are based on hierarchical agglomerative clustering (HAC). 
For instance, in \cite{Melnykov11, Maitra09, Meila98}, HAC (with different distance measures)
is used to obtain mean vectors.
Since HAC is generally very slow, it is usually only executed on a random sample \cite{Meila98}.
However, the size of any reasonable sample depends on the size of the smallest optimal component.
Moreover, it is often outperformed by other methods (e.g. \cite{Melnykov11,Meila98}).
Another approach using HAC is presented in \cite{Maitra09}.
It aims at finding the best local modes of the data set in a reduced $m^*$-dimensional space and applies HAC only on these modes. 
However, this method is time-consuming and the choice of $m^*$ is crucial \cite[p.~5,13]{Maitra09}.
Moreover, in \cite{Maitra10} it is outperformed by simple random methods. 

\cite{Melnykov11} presents a density based approach which not only determines an initial solution but also the number of components.
It initializes the means by points which have a ``high concentration'' of neighbors. 
To this end, the size $m$ of the neighborhood of a point (i.e., the minimum number of points in a cluster) has to be fixed in advance. 
In our experiments, we found that the performance crucially depends on the choice of $m$.
Hence, we ignore this method in this paper.

In \cite{Verbeek03}, a greedy algorithm is presented which constructs a sequence of mixture models with $1$ through $K$ components. 
Given a model $\theta_k$ with $k$ components, it constructs several new candidates with $k+1$ components. 
Each candidate is constructed by adding a new component to $\theta_k$ and executing the EM algorithm.
Hence, this method is only useful if several values of $K$ need to be considered.

In \cite{Kwedlo2015} a modification of the Gonzalez algorithm for GMMs is presented. 
Furthermore, there are some practical applications using the $K$-means++ algorithm for the initialization of GMMs (e.g., in \cite{Krueger10} GMMs are used for speech recognition).
Additional initialization methods can be found e.g. in \cite{Maitra09, Thiesson98, Fayyad98, Biernacki04}.

\subsection{Our Contribution}\label{sec:intro:contribution}
Clearly, there is no way to determine the \emph{best} initialization algorithm that outperforms all other algorithms on all instances.
The performance of an initialization depends on the given data and the allowed computational cost.
Nonetheless, the initializations presented so far (except the simple random initializations) face mainly two problems:
Firstly, they are rather complex and time consuming. 
Secondly, the choice of hyperparameters is crucial for the outcome.

In this paper, we present new methods that are fast and {do not require choosing sensitive hyperparameters}.
These methods can be seen as adaptions of the $K$-means++ algorithm \cite{ArthurV07} and the Gonzalez algorithm \cite{Gonzalez85} and as an extension of the initial work in \cite{Kwedlo2015, Krueger10}.
We present experiments indicating the superiority of our methods compared to a large number of alternative methods. 

\section{Baseline Algorithms}\label{sec:baselines}
The most widely used initializations start by choosing $K$ data points:
\begin{description}
  \item[\texttt{Unif}] draws $K$ points independently and uniformly at random from $X$.
  \item[\texttt{HAC}] computes a uniform sample $S$ of size $s\cdot |X|$ of the input set $X$ and executes hierarchical clustering with average linkage cost on $S$.
   \item[\texttt{G}]  executes the algorithm given in \cite{Gonzalez85},
 which yields a 2-approximation for the discrete radius $K$-clustering problem. 
 Iteratively, it chooses the point with the largest Euclidean distance from the already chosen points.
  \item[\texttt{KM++}]  executes the $K$-Means++ algorithm \cite{ArthurV07}, which has been designed for the $K$-means problem.
In each round, \texttt{KM++} samples a data point (i.e. the next mean) from the given data set $X$ with probability proportional to its $K$-means cost (with respect to the points chosen so far). 
In expectation, the resulting $K$-means costs are in $\mathcal{O}(\log(K) \cdot \mbox{opt})$.
\texttt{KM++} is particularly interesting since the $K$-means algorithm is a special case of the  EM algorithm \cite{Bishop06}.
\end{description}
Then, given $K$ data points, Alg.~\ref{algo:means2gmm} is used to create a GMM, which is then the initial solution that is fed to the EM algorithm. 
\begin{algorithm}[htb]
\begin{algorithmic}[1]
    \State  Derive a partition $\{C_1,\ldots,C_k\}$ by assigning each $x\in X$ to a closest point in $\mathcal{C}$.
    \For{$l=1,\ldots,k$}
      \State Set $\mu_l \coloneqq \nicefrac{1}{|C_l|}\sum_{x\in C_l} x$, $ w_l \coloneqq \nicefrac{|C_l|}{|X|} $,  $\Sigma_l \coloneqq \nicefrac{1}{|C_l|}\sum_{x\in C_l} (x-\mu_l)(x-\mu_l)^T$. \label{algo:means2gmm:update}
      \State If $\Sigma_l$ is not positive definite, set $\Sigma_l\coloneqq \nicefrac{1}{(D\cdot \vert C_l\vert) }\sum_{x\in C_l} \Vert x - \mu_l \Vert^2 \cdot I_D$
      \State If $\Sigma_l$ is still not positive definite, set $\Sigma_l \coloneqq I_D$
    \EndFor
    \State \Return $\theta = \{(w_l,\mu_l,\Sigma_l)\}_{l=1,\ldots,k}$.
\end{algorithmic}
\caption{\texttt{Means2GMM}($X\subset\R^D$, $\mathcal{C}\subset\R^D$, $|\mathcal{C}|=k$)}
\label{algo:means2gmm}
\end{algorithm}

A popular alternative is to apply the $K$-means algorithm with the chosen data points before executing Alg.~\ref{algo:means2gmm}. 
The main idea behind this is that starting the EM algorithm with a coarse initial solution 
(where e.g. not all clusters are covered well) 
might impose a high risk of getting stuck at a poor local minimum.
To avoid this problem, one first runs a different algorithm that optimizes a function similar to the likelihood, i.e. the $K$-means costs (cf. \cite[p.~427, p.~443]{Bishop06}). 
We refer to the $K$-means algorithm as an \emph{intermediate} algorithm and indicate its use by the postfix ``$_{km}$''.

\section{Adaptive Seeding for GMMs}\label{sec:initial:adaptive}
Our new adaptive methods construct a sequence of models with $k=1$ through $k=K$ components adaptively.
Given a $(k-1)$-GMM $\theta_{k-1}$, our methods try to choose a point from the data set that is not described well by the given $\theta_{k-1}$ and which is hopefully a good representative of a component of an optimal $k$-GMM. 
The idea behind is that this point can lead us to a significant refinement of $\theta_{k-1}$.

\paragraph{Choosing a Point.}
The negative log-likelihood of a point $x\in\mathbb{R}^D$, given the GMM $\theta_{k-1}$, measures how well $x$ is described by $\theta_{k-1}$.\footnote{
The (inverse) pdf is unsuited due to the exponential behavior (over-/underflows).}
Unfortunately, it may take negative values and 
does not scale with the data set and the GMM\footnote{
even wrt. a single Gaussian, i.e. $\log\cN(c\cdot x|c\cdot\mu,c^2\cdot\Sigma)=\log\cN(x|\mu,\Sigma)-D \ln(c)$}.
This also applies to the minimum component-wise negative log-likelihood
\begin{align*}
 \textstyle \min\left\{ - \log\left((2\pi)^{D/2}\vert\Sigma_l\vert^{1/2}\right) 
 + \frac{1}{2}(x-\mu_l)^T\Sigma_l^{-1}(x-\mu_l)\,\mid\, (w_l,\mu_l,\Sigma_l)\in\theta_{k-1} \right\} \ ,
\end{align*}
due to the first summand. 
Hence, we use  the {minimum Mahalanobis distance}
\[ \textstyle m(x| \theta_{k-1})\coloneqq \min\left\{ (x-\mu_l)^{T}\Sigma_l^{-1}(x-\mu_l)\,\mid\, (w_l,\mu_l,\Sigma_l)\in\theta_{k-1} \right\}\ . \]
Our first method chooses the point $x\in X$ maximizing $m(x|\theta_{k-1})$.
Since these points are more likely to be outliers, we also consider choosing a point only from a uniform sample of $X$, which is chosen in advance (cf. Alg.~\ref{algo:gonzalez2GMM}).

Our second method chooses point $x\in X$ with probability $\propto m(x|\theta_{k-1})$ (cf. Alg.~\ref{algo:adaptiveMeans2GMM}). 
In order to reduce the probability to choose an outlier, we also consider adding an $\alpha$ portion of uniform distribution, i.e. drawing $x$ with probability
\[ \textstyle m_{{\alpha}}(x|\theta_{k-1}) \coloneqq  \alpha \cdot {m(x|\theta_{k-1})}/{\sum_{y\in X} m(y|\theta_{k-1})} + (1-\alpha)\cdot{1}/{\vert X\vert}\ . \]

\paragraph{Constructing a GMM.}
Then, given a point $x\in X$ and the means of $\theta_{k-1}$, we construct a $k$-GMM. 
In our first experiments, we used Alg.~\ref{algo:means2gmm} to construct a $k$-GMM. 
However, it turned out that estimating only spherical covariance matrices (with variable variances) yields a better performance than estimating full covariance matrices. 
We assume that this is due to the fact that $\theta_{k-1}$ is only a very coarse estimate of $(k-1)$-components of an optimal $k$-GMM.
Formally, we replace the covariance update in Line~\ref{algo:means2gmm:update} of Alg.~\ref{algo:means2gmm} by 
$\Sigma_l = \nicefrac{1}{(D\cdot \vert C_l\vert) }\sum_{x\in C_l} \Vert x - \mu_l \Vert^2 \cdot I_D$. 
We denote this version of Alg.~\ref{algo:means2gmm} as \texttt{Means2SphGMM}. 
Given the resulting $k$-GMM, our methods then again choose a new point from $X$ as already described above.

\paragraph{Intermediate Algorithm.}
Recall that some baselines use the $K$-means algorithm as an intermediate algorithm (cf. Sec.~\ref{sec:baselines}). 
Since we do not only construct means but GMMs, we apply a hard-clustering variant of the EM algorithm, i.e. the {Classification EM algorithm} (CEM) \cite{Celeux92}, and let it only estimate spherical covariances. 
We indicate its use by the postfix "$_{cem}$" .

Alg.~\ref{algo:adaptiveMeans2GMM} and Alg.~\ref{algo:gonzalez2GMM} summarize our methods.
Note that we do not optimize the hyperparameters $\alpha$ and $s$ in our experiments.

\begin{figure*}[ttt!]
 \begin{minipage}[t]{.52\textwidth}
\begin{algorithm}[H]
  \begin{algorithmic}[1]
    \Require{$X\subset\R^D,K\in\N,s\in(0,1]$}
    \State  $\theta_1 \coloneqq  $ optimal 1-MLE wrt. $X$
    \State If $s<1$, let $S$ be a uniform sample of $X$ of size $\lceil s\cdot|X| \rceil$. Otherwise, set $S=X$. 
    \For{$k=2,\ldots,K$}
      \State $p \coloneqq  \arg\max_{x\in S} m(x|\theta_{k-1})$
      \State $M_k \coloneqq \{\mu|(\cdot,\mu,\cdot)\in \theta_{k-1}\}\cup\{p\}$ 
      \State $\theta_k \coloneqq  \texttt{Means2SphGMM}(X, M_k)$
    \EndFor
    \State (optional) Run CEM algorithm
    \State \Return $\theta_K$
  \end{algorithmic}
  \caption{\texttt{SphericalGonzalez (SG)}}
  \label{algo:gonzalez2GMM}
\end{algorithm}
 \end{minipage}
 \hfill
 \begin{minipage}[t]{.46\textwidth}
\begin{algorithm}[H]
  \begin{algorithmic}[1]
     \Require{$X\subset\R^D,K\in\N,\alpha\in[0,1]$}
     \State  $\theta_1 \coloneqq  $ optimal 1-MLE wrt. $X$
      \For{$k=2,\ldots,K$}
	\State Draw $p$ from $X$ with probability $m_{{\alpha}}(p|\theta_{k-1})$.
	\State $M_k \coloneqq  \{\mu|(\cdot,\mu,\cdot)\in \theta_{k-1}\}\cup\{p\}$ 
	\State $\theta_k \coloneqq  \texttt{Means2SphGMM}(X, M_k)$
      \EndFor
      \State (optional) Run CEM algorithm
      \State \Return $\theta_K$
  \end{algorithmic}
  \caption{\texttt{Adaptive (Ad)}}
  \label{algo:adaptiveMeans2GMM}
\end{algorithm}
 \end{minipage}
 \end{figure*}

\paragraph{Comparison to Baselines.}
Our adaptive initializations can be seen as adaptions of the \texttt{Gonzalez} and \texttt{Kmeans++} algorithm. 
Simply speaking, these methods assume that each component is represented by a Gaussian with \emph{the same fixed spherical covariance} matrix and fixed uniform weights. 
In contrast, our goal is to estimate also the covariance matrices \emph{adaptively}. 
Furthermore, in \cite{Kwedlo2015} another adaption of the Gonzalez algorithm is presented, which we denote by \texttt{KwedlosGonzalez (KG)}. 
Unlike our method, it chooses weights and covariance matrices randomly and independently of the means (and of each other).

\section{Experiments}\label{sec:experiments}
We evaluated all presented methods with respect to artificial as well as real world data sets.
Our implementation as well as the complete results are available at \cite{SupplementalMaterial}.
Due to space limitations, we omit the results of those algorithms that are consistently outperformed by others. 
These results are available at \cite{SupplementalMaterial} as well.

\paragraph{Quality Measure.}
Recall that the goal of our paper (and the EM algorithm) is to find a maximum likelihood estimate (MLE). 
Thus, the \emph{likelihood} is not only the common but also the appropriate way of evaluating our methods.

Other measures need to be treated with caution:  
Some authors consider their methods only with respect to \emph{some specific tasks} where fitting a GMM to some data is part of some framework. 
Hence, any observed effects might be due to several reasons (i.e. correlations). 
In particular, GMMs are often compared with respect to certain \emph{classifications}. 
As already pointed out by \cite{FGK}, the class labels of real world data sets do not necessarily correspond to the structure of an MLE.
The same holds for data sets and classifications generated according to some GMM.
Moreover, a \emph{cross-validation}, that examines whether methods over-fit models to training data,  is not reasonable, since our methods do \emph{not} aim at finding a model that does not fit too well to the given data set. 
Finally, one should not generate data sets according to some \emph{``ground truth`` GMM} $\theta_{gt}$ and compare GMMs with $\theta_{gt}$ because in many cases (e.g. small $|X|$) one cannot expect $\theta_{gt}$ to be a good surrogate of the MLE.

\paragraph{Setup.}
Recall that in Alg.~\ref{algo:gonzalez2GMM} and Alg.~\ref{algo:adaptiveMeans2GMM} hyperparameters $\alpha$ and $s$ are used. 
We do not optimize them, but test reasonable values, i.e. $\alpha\in\{0.5,1\}$ and $s\in\{0.1,1\}$.
We execute each method with 30 different seeds. 
On the basis of some initial experiments, we decided to execute the intermediate algorithms for 25 rounds and the EM algorithm for 50 rounds.
If only the EM algorithm is applied, then we execute it for 75 rounds.

\subsection{Artificial Data Sets}\label{sec:experiments-generated}
\subsubsection{Data Generation.}
We create 192 test sets, each containing 30 data sets that share certain characteristics. \cite{SupplementalMaterial}.  
For each data set, we first create a GMM\footnote{
As explained before, our goal is \emph{not} to identify these GMMs. 
} at random but control the following properties: 
First, the components of a GMM can either be spherical or elliptical. 
We describe the eccentricity of $\Sigma_k$ by $e_k = \frac{\max_d \lambda_{kd}}{\min_d \lambda_{kd}}$, where $\lambda_{kd}^2$ denotes the $d$-th eigenvalue of $\Sigma_k$.
Second, components can have different sizes, in terms of the smallest eigenvalue of the corresponding covariance matrices.
Third, components have different weights. 

Fourth, components can overlap more or less. 
Following \cite{Dasgupta99}, we define the separation parameter
$ c_\theta = \min_{\ l,k} {\Vert \mu_l - \mu_k \Vert }/{\sqrt{\max\left\{\mbox{trace}(\Sigma_l),\mbox{trace}(\Sigma_k)\right\}}}$. 
\begin{wrapfigure}{r}{0.5\textwidth}
\vspace{-5mm}
\centering
\subfloat[$c_\theta=0.5$]{
\includegraphics[height=.14\textwidth]{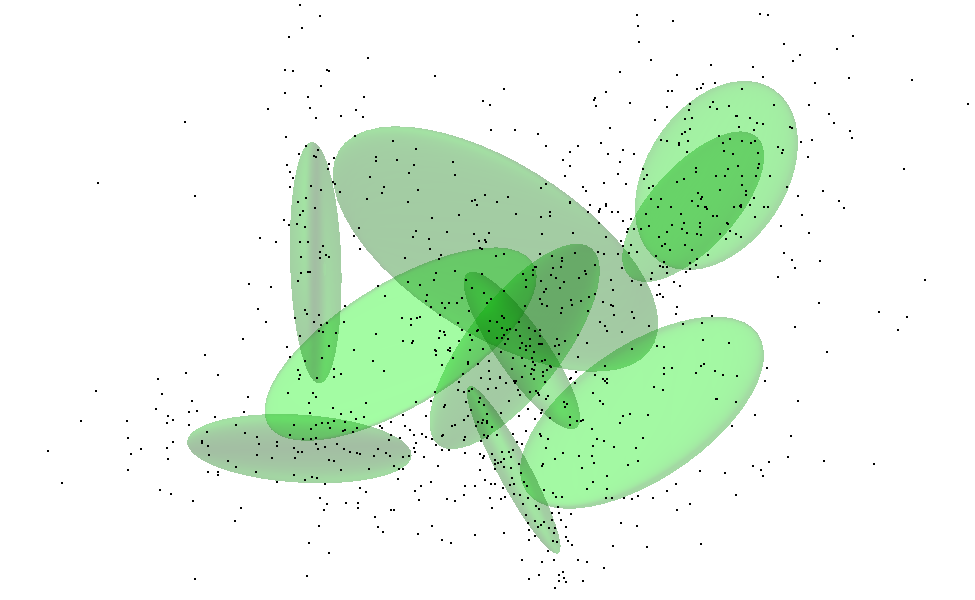}
\label{fig:sep05}}
\subfloat[$c_\theta=1$]{
\includegraphics[height=.14\textwidth]{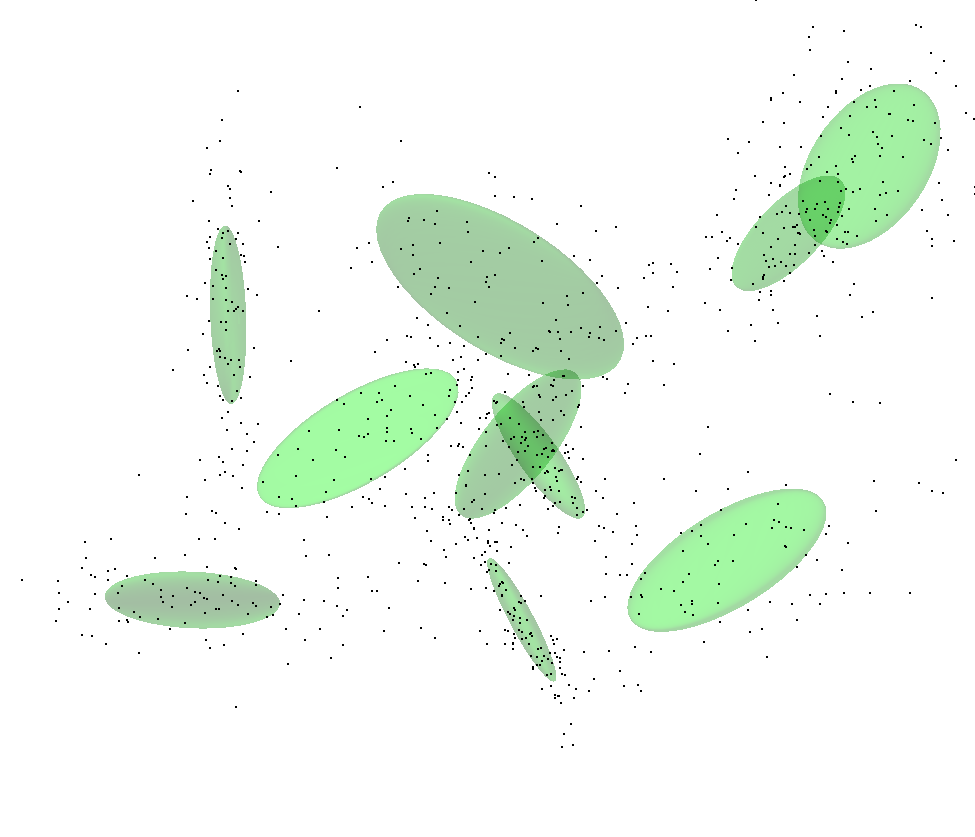}
\label{fig:sep1}}
\subfloat[ $c_\theta=2$]{
\includegraphics[height=.14\textwidth]{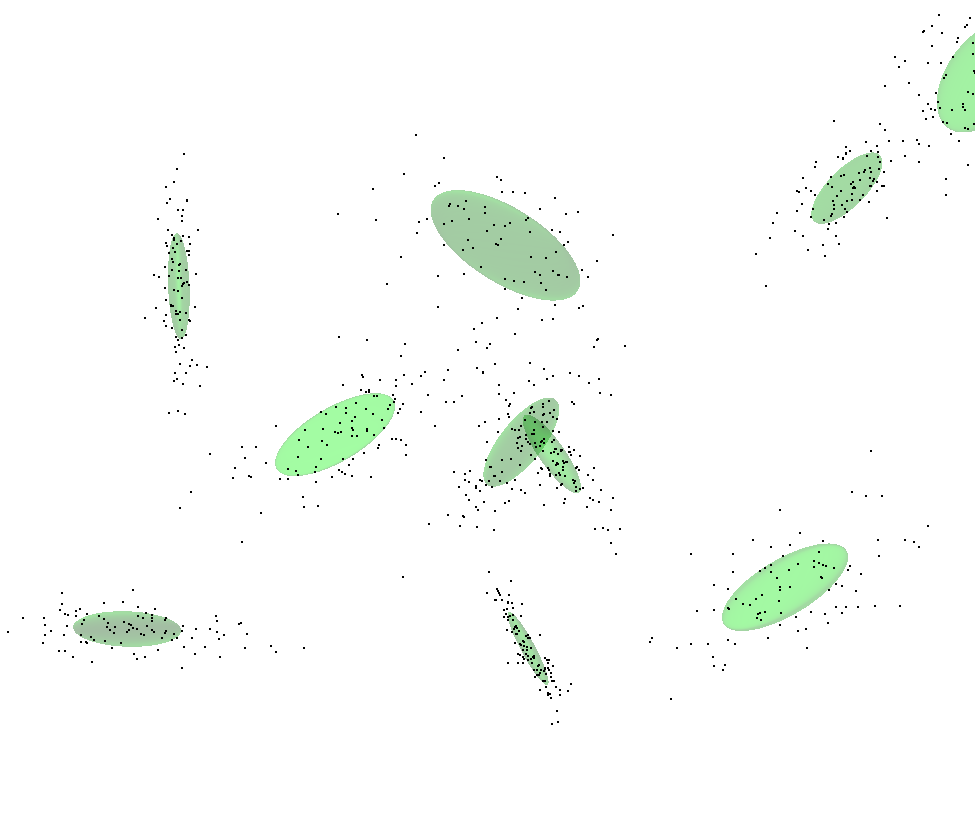}
\label{fig:sep2}}

\subfloat[$c_\theta=0.5$]{
\includegraphics[width=.14\textwidth]{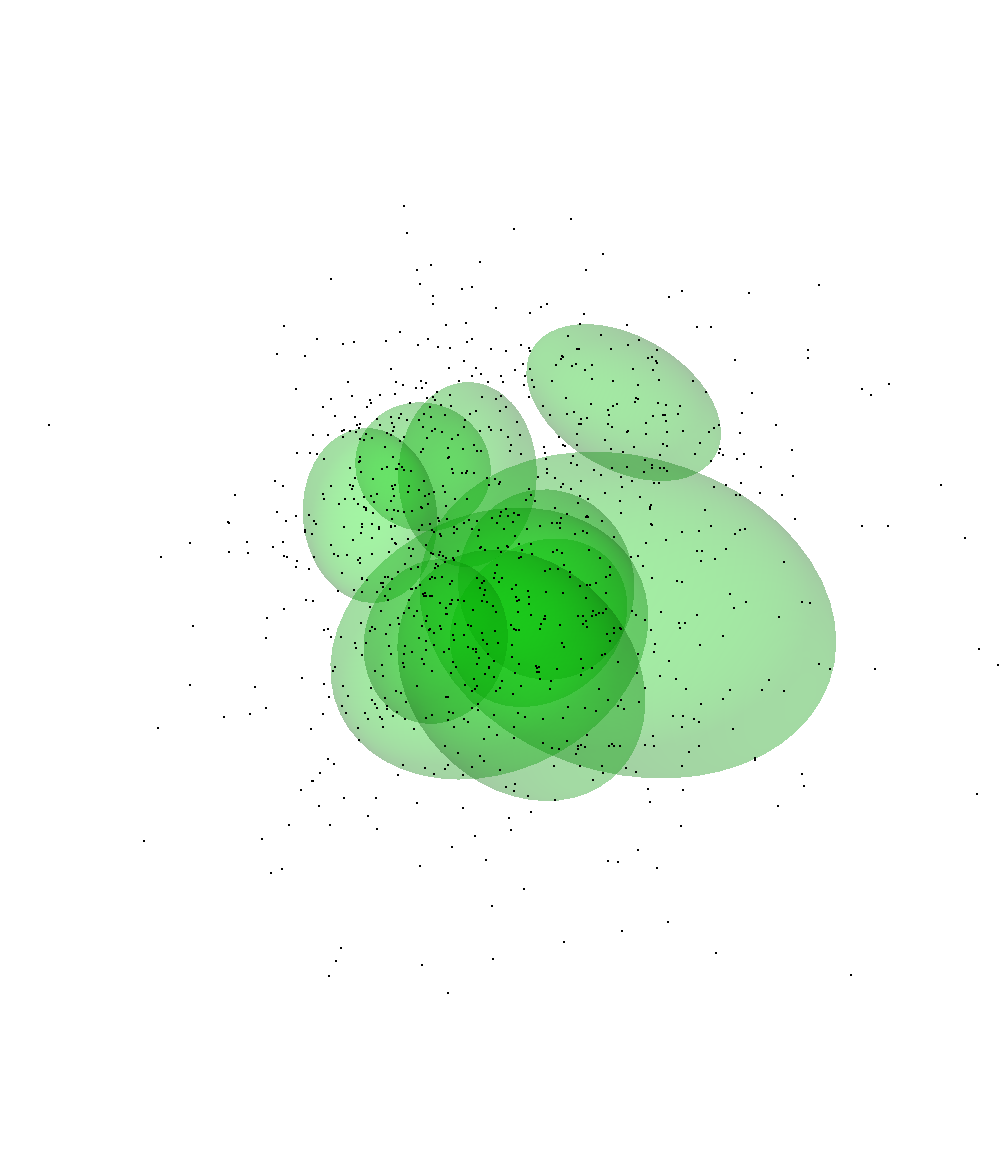}
\label{fig:d10sep05}}
\subfloat[$c_\theta=1$]{
\includegraphics[width=.14\textwidth]{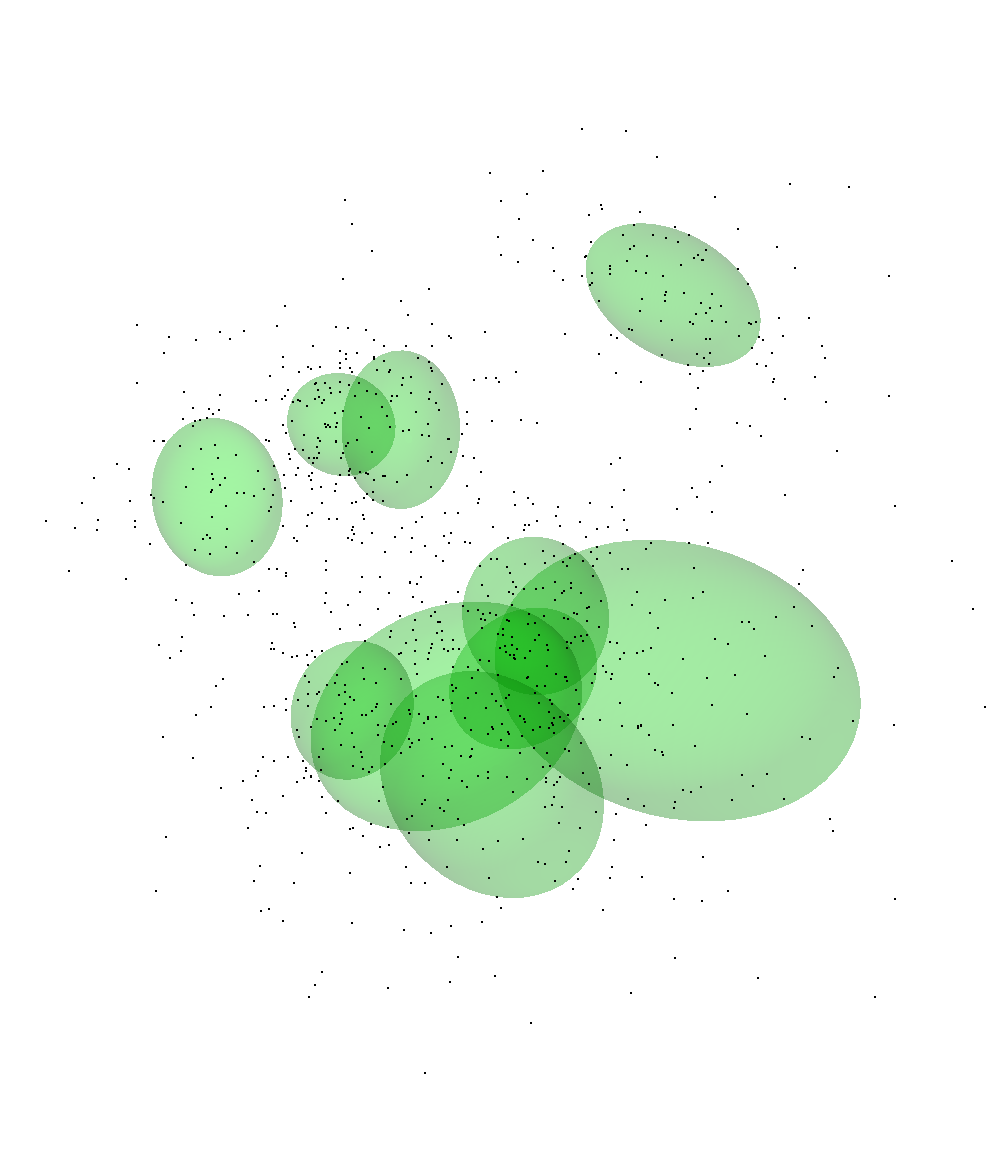}
\label{fig:d10sep1}}
\subfloat[$c_\theta=2$]{
\includegraphics[width=.14\textwidth]{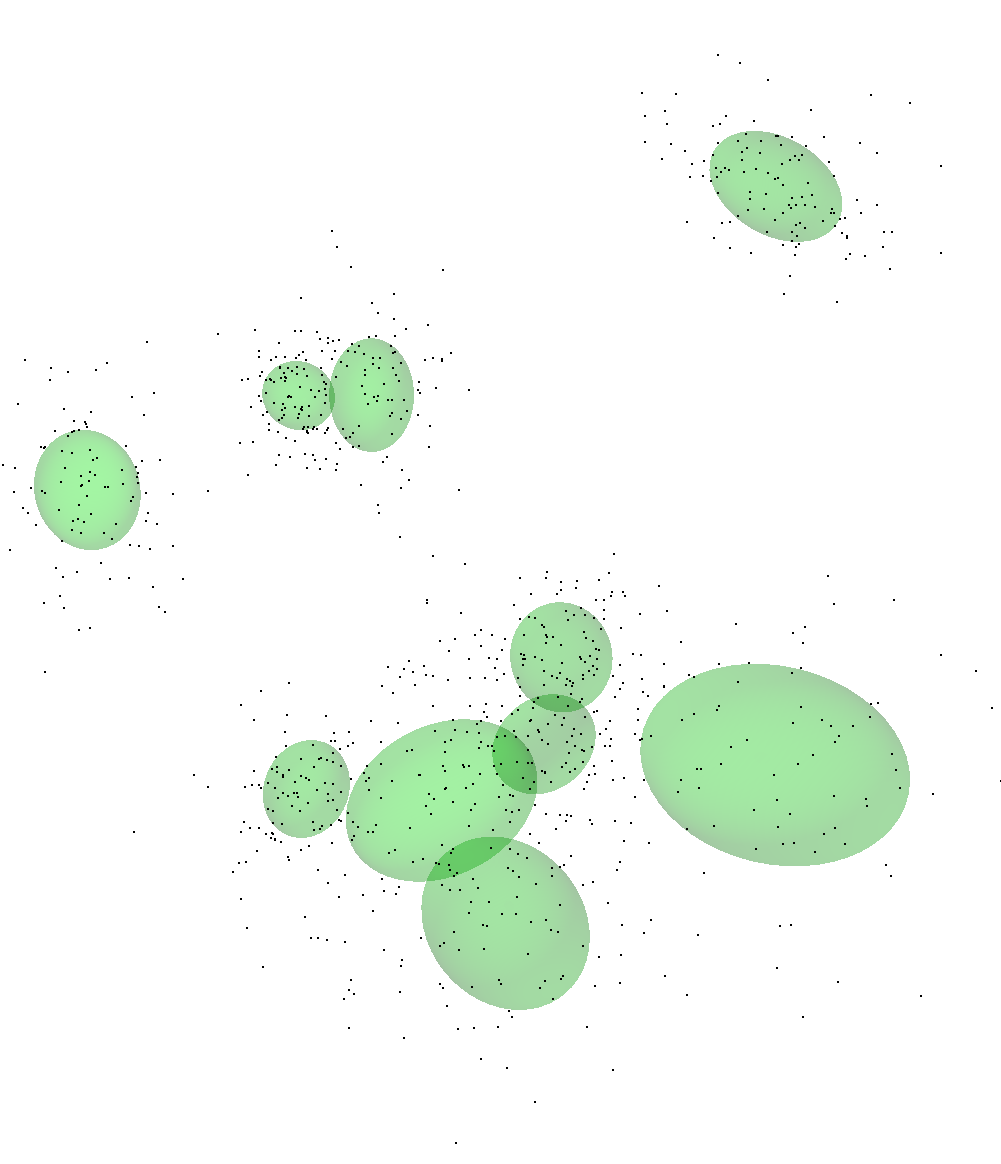}
\label{fig:d10sep2}}
\caption{ Examples for different separation parameters. 
Figures show orthogonal projections to random plane. 
Data sets in \eqref{fig:sep05}--\eqref{fig:sep2} have $D=3$. 
\eqref{fig:d10sep05}--\eqref{fig:d10sep2} have $D=10$. 
}
\label{fig:gendata}
\end{wrapfigure}
In {high dimension} $D\gg 1$, $c_\theta=2$ corresponds to almost completely separated clusters (i.e. points generated by the same component), while $c_\theta\in\{0.5,1\}$ indicates a slight but still negligible overlap \cite{Dasgupta00}. 
However, in {small dimension}, $c_\theta\in\{0.5,1\}$  corresponds to significant overlaps between clusters, while  $c_\theta=2$  implies rather separated clusters (cf. Fig.~\ref{fig:gendata}). 

We generate random GMMs as follows.
Initially, we draw means uniformly at random from a cube with a fixed side length. 
For the weights, we fix some $c_w\geq 0$, construct a set of weights $\{ 2^{c_w\cdot i}/\sum_{j=1}^K 2^{c_w\cdot j} \}_{i=1,\ldots,K}$ and assign these weights randomly.
To control the sizes and the eccentricity, we fix the minimum and maximum eigenvalue and draw the remaining values uniformly at random from the interval.
Then, we set  $\Sigma_k = Q^T\mbox{diag}(\lambda_{k1}^2,\ldots,\lambda_{kD}^2)Q$ for a random $Q\in \text{SO}(D)$. 
Finally, the means are scaled as to fit the predefined separation parameter.
Given the resulting GMM $\theta$, we first draw some points according to $\theta$. 
Then, we construct a bounding box, elongate its side lengths by a factor 1.2, and draw noise points uniformly at random from the resized box.

We created a test set (i.e. 30 data sets) for each combination of the following parameters:
$K=20$, $|X|\in\{1\,000,\ 5000\}$, $D\in\{3,10\}$, $c_\theta\in\{0.5,\ 1,\ 2\}$, $c_w\in\{0.1, 1\}$, different combinations of size and eccentricity ( i.e., equal size and $e_k=10$, equal size and $e_k\in [1,10]$, different size and $e_k=1$, different size and $e_k \in [1,10]$),
and   without or with 10\% noise points.

\subsubsection{Evaluation Method.}
We consider the \emph{initial solutions} produced by the initialization (possibly followed by an intermediate algorithm) and the \emph{final solutions} obtained by running the EM algorithm afterwards.
For each data set, we compute the average log-likelihood of the initial and final solution, respectively.
Based on these averages, we create rankings of the algorithms\footnote{Averaging the (average) log-likelihood values over different data sets is not meaningful since the optimal log-likelihoods may deviate significantly. }.
Then, we compute the average rank (and standard deviation of the rank) of each algorithm over all datasets matching certain properties. 

\subsubsection{Results.}
In general, we observe that one should use an intermediate algorithm before applying the EM algorithm. 
Thus, we omit the results of some methods \cite{SupplementalMaterial}. 
\paragraph{Data without Noise.}
For these rather simple data sets, there is no method that constantly outperforms \emph{all others}.
Nonetheless, it is always one of our adaptive or the \texttt{G}$_{km}$ initialization that performs best.

The results depcited in Tab.~\ref{exp:generated:tab:d10:without:weighted} and \ref{exp:generated:tab:d10:without:unweighted} suggest that, 
regardless of the weights, the performance is determined by the separation.
Furthermore, a good initial solution does not imply a good final solution. 
Given overlap ($c_{\theta}=0.5$) or moderate separation ($c_{\theta}=1$), \texttt{SG}$_{cem}(s=1)$ and \texttt{Ad}$_{cem}(\alpha=1)$  work best, even though their initial solutions have low average ranks compared to \texttt{KM++}$_{km}$. 
Given higher separation ($c_{\theta}=2$), we expect it to be easier to identify clusters and that skewed covariance matrices do not matter much if means are assigned properly in the first place. 
Indeed, the simple \texttt{G}$_{km}$ and \texttt{KG} do the trick.
\begin{table*}[htb]
  \centering
  \caption{Average ranks ($\pm$ std.dev.) for generated data with $K=20$, $|X|=1\,000$, $D=10$, \textbf{different weights}, and \textbf{without noise.}}
  \label{exp:generated:tab:d10:without:weighted}
    \begin{tabular}{l|ll|ll|ll}
          &  \multicolumn{2}{c|}{\bf separation $c_{\theta}=0.5$}     &  \multicolumn{2}{c|}{\bf separation $c_{\theta}=1$}     &  \multicolumn{2}{c}{\bf separation $c_{\theta}=2$}  \\
          & \multicolumn{1}{c}{initial} & \multicolumn{1}{c|}{final}  & \multicolumn{1}{c}{initial} & \multicolumn{1}{c|}{final} & \multicolumn{1}{c}{initial} & \multicolumn{1}{c}{final} \\ 
\texttt{SG}$\left(s=\frac{1}{10}\right)$ & 7.53$\pm$1.08 & 3.58$\pm$1.93 & 7.29$\pm$0.86 & 5.08$\pm$2.95 & 7.14$\pm$0.57 & 7.28$\pm$2.31 \\
\texttt{SG}$(s=1)$ & 8.00$\pm$1.52 & \blue 3.26$\pm$2.27 & 8.77$\pm$0.68 & 5.53$\pm$3.42 & 8.72$\pm$0.53 & 7.44$\pm$2.60 \\
 \texttt{KG}$(s=1)$ & 10.00$\pm$0.00 & 9.75$\pm$0.72 & 10.00$\pm$0.00 & 7.68$\pm$2.52 & 10.00$\pm$0.00 & \blue 2.38$\pm$1.34 \\
\texttt{Unif}$_{{km}}$ & 1.56$\pm$0.74 & 8.39$\pm$1.15 & 2.19$\pm$0.61 & 7.46$\pm$1.88 & 2.98$\pm$0.13 & 7.08$\pm$2.14 \\
\texttt{G}$_{{km}}$  & 3.34$\pm$1.33 & 6.03$\pm$2.32 & 2.38$\pm$0.86 & 5.16$\pm$2.79 & 1.23$\pm$0.46 & \blue 1.99$\pm$1.31 \\
\texttt{KM++}$_{{km}}$  & 1.85$\pm$0.64 & 7.87$\pm$1.14 & 1.43$\pm$0.64 & 6.22$\pm$2.60 & 1.78$\pm$0.41 & 3.75$\pm$2.25 \\
\texttt{SG}$\left(s=\frac{1}{10}\right)_{{cem}}$  & 6.15$\pm$0.60 & 3.95$\pm$1.24 & 6.30$\pm$0.68 & 4.71$\pm$2.54 & 6.10$\pm$0.40 & 6.65$\pm$2.13 \\
\texttt{SG}$(s=1)_{{cem}}$  & 6.47$\pm$1.51 & \blue 3.20$\pm$2.48 & 7.32$\pm$1.26 & 5.12$\pm$3.31 & 8.03$\pm$0.61 & 7.80$\pm$2.61 \\
 \texttt{Ad}$(\alpha=1)_{{cem}}$ & 5.16$\pm$1.84 & 4.31$\pm$1.77 & 4.59$\pm$0.64 & \blue 3.88$\pm$1.75 & 4.47$\pm$0.50 & 5.06$\pm$1.40 \\
\texttt{Ad}$\left(\alpha=\frac{1}{2}\right)_{{cem}}$ & 4.93$\pm$2.11 & 4.67$\pm$1.93 & 4.72$\pm$0.80 & \blue 4.15$\pm$1.77 & 4.53$\pm$0.50 & 5.58$\pm$1.71 
    \end{tabular}%
\end{table*}%
\begin{table*}[htb]
  \centering
  \caption{Average ranks ($\pm$ std.dev.) for generated data with $K=20$, $|X|=1\,000$, dimension $D=10$, \textbf{equal weights}, and \textbf{without noise}.}
  \label{exp:generated:tab:d10:without:unweighted}
    \begin{tabular}{l|cc|cc|cc}
          &  \multicolumn{2}{c|}{\bf separation $c_{\theta}=0.5$}     &  \multicolumn{2}{c|}{\bf separation $c_{\theta}=1$}     &  \multicolumn{2}{c}{\bf separation $c_{\theta}=2$}  \\
          & \multicolumn{1}{c}{initial} & \multicolumn{1}{c|}{final}  & \multicolumn{1}{c}{initial} & \multicolumn{1}{c|}{final} & \multicolumn{1}{c}{initial} & \multicolumn{1}{c}{final} \\ 
\texttt{SG}$\left(s=\frac{1}{10}\right)$ & 7.58$\pm$0.98 & 3.98$\pm$1.87 & 7.36$\pm$0.73 & 5.02$\pm$2.97 & 7.08$\pm$0.41 & 7.35$\pm$2.24 \\
\texttt{SG}$(s=1)$ & 8.11$\pm$1.53 & 3.62$\pm$2.58 & 8.67$\pm$0.85 & 5.67$\pm$3.20 & 8.79$\pm$0.43 & 7.77$\pm$2.59 \\
 \texttt{KG}$(s=1)$ & 10.00$\pm$0.00 & 9.54$\pm$0.96 & 10.00$\pm$0.00 & 7.97$\pm$2.40 & 10.00$\pm$0.00 & \blue 2.38$\pm$1.23 \\
\texttt{Unif}$_{{km}}$ & 1.44$\pm$0.70 & 8.36$\pm$1.25 & 2.14$\pm$0.61 & 7.28$\pm$1.83 & 2.98$\pm$0.16 & 7.00$\pm$1.98 \\
\texttt{G}$_{{km}}$  & 3.39$\pm$1.34 & 6.12$\pm$2.17 & 2.53$\pm$0.83 & 6.04$\pm$2.76 & 1.27$\pm$0.50 & \blue 1.82$\pm$1.08 \\
\texttt{KM++}$_{{km}}$  & 1.91$\pm$0.55 & 7.82$\pm$1.30 & 1.35$\pm$0.56 & 5.89$\pm$2.77 & 1.75$\pm$0.43 & 3.77$\pm$2.50 \\
\texttt{SG}$\left(s=\frac{1}{10}\right)_{{cem}}$  & 6.18$\pm$0.62 & \blue 3.58$\pm$1.31 & 6.44$\pm$0.87 & 4.26$\pm$2.43 & 6.02$\pm$0.13 & 6.55$\pm$1.95 \\
\texttt{SG}$(s=1)_{{cem}}$  & 6.49$\pm$1.44 & \blue 3.17$\pm$2.73 & 7.35$\pm$1.13 & 5.30$\pm$3.23 & 8.12$\pm$0.45 & 8.04$\pm$2.37 \\
 \texttt{Ad}$(\alpha=1)_{{cem}}$ & 5.03$\pm$1.75 & 4.12$\pm$1.79 & 4.49$\pm$0.64 & \blue 3.48$\pm$1.61 & 4.42$\pm$0.50 & 4.90$\pm$1.35 \\
\texttt{Ad}$\left(\alpha=\frac{1}{2}\right)_{{cem}}$ & 4.87$\pm$1.99 & 4.69$\pm$1.91 & 4.67$\pm$0.65 & \blue 4.09$\pm$1.73 & 4.58$\pm$0.50 & 5.42$\pm$1.38 
    \end{tabular}%
\end{table*}%

Tab.~\ref{exp:generated:tab:d10:without:ecc} shows that \texttt{Ad}$_{cem}(\alpha=1)$ works well for elliptical data, while \texttt{G}$_{km}$ should be chosen for spherical data.
Recall that there are no noise points yet.
We expect that the performance of \texttt{G}$_{km}$ degenerates in the presence of noise since it is prone to choose outliers.
Overall, given data sets without noise, \texttt{Ad}$_{cem}(\alpha=1)$ performs best. 
\begin{table*}[htb]
  \centering
  \caption{Average ranks ($\pm$ std.dev.) for generated data with $K=20$, $|X|=1\,000$, dimension $D=10$, and \textbf{without noise}.
  Only final solutions. }
  \label{exp:generated:tab:d10:without:ecc}
\begin{tabular}{l|ccc|ccc}
      & \multicolumn{3}{c|}{\bf equal weights} & \multicolumn{3}{c}{\bf different weights} \\
      & \bf{spherical} & \bf{elliptical} & \bf{both}   & \bf{spherical} & \bf{elliptical} & \bf{both} \\
{\texttt{SG}$\left(s=\frac{1}{10}\right)$} & 6.13$\pm$2.63 & 5.22$\pm$2.80 & 5.45$\pm$2.78 & 6.03$\pm$2.65 & 5.07$\pm$2.90 & 5.31$\pm$2.87 \\
{\texttt{SG}$(s=1)$} & 6.64$\pm$2.99 & 5.37$\pm$3.30 & 5.69$\pm$3.27 & 6.04$\pm$3.22 & 5.20$\pm$3.28 & 5.41$\pm$3.28 \\
{\texttt{KG}$(s=1)$} & 6.78$\pm$3.19 & 6.58$\pm$3.59 & 6.63$\pm$3.49 & 6.81$\pm$3.20 & 6.53$\pm$3.65 & 6.60$\pm$3.54 \\
{\texttt{Unif}$_{{km}}$} & 7.67$\pm$1.48 & 7.50$\pm$1.91 & 7.54$\pm$1.81 & 7.64$\pm$1.64 & 7.64$\pm$1.92 & 7.64$\pm$1.85 \\
{\texttt{G}$_{{km}}$ } & \blue 3.03$\pm$2.26 & 5.20$\pm$2.92 & \blue 4.66$\pm$2.92 & \blue 2.83$\pm$2.33 & \blue 4.91$\pm$2.78 & \blue 4.39$\pm$2.82 \\
{\texttt{KM++}$_{{km}}$ } & 5.44$\pm$2.60 & 5.96$\pm$2.87 & 5.83$\pm$2.81 & 5.57$\pm$2.66 & 6.07$\pm$2.69 & 5.95$\pm$2.69 \\
{\texttt{SG}$\left(s=\frac{1}{10}\right)_{{cem}}$ } & 4.77$\pm$2.61 & \blue 4.81$\pm$2.23 & 4.80$\pm$2.33 & 5.68$\pm$2.19 & \blue 4.91$\pm$2.35 & 5.10$\pm$2.33 \\
{\texttt{SG}$(s=1)_{{cem}}$ } & 6.53$\pm$3.16 & 5.16$\pm$3.46 & 5.51$\pm$3.43 & 6.07$\pm$3.30 & 5.14$\pm$3.40 & 5.37$\pm$3.39 \\
{\texttt{Ad}$(\alpha=1)_{{cem}}$} & \blue 3.62$\pm$1.61 & \blue 4.34$\pm$1.69 & \blue 4.16$\pm$1.69 & \blue 4.02$\pm$1.82 & \blue 4.55$\pm$1.66 & \blue 4.42$\pm$1.71 \\
{\texttt{Ad}$\left(\alpha=\frac{1}{2}\right)_{{cem}}$} & 4.38$\pm$1.78 & 4.86$\pm$1.75 & 4.74$\pm$1.77 & 4.30$\pm$1.86 & 4.97$\pm$1.88 & 4.80$\pm$1.89
\end{tabular}%
\end{table*}

\paragraph{Noisy Data.}
When introducing noise, our adaptive methods are still among the best methods, while the performance of some others degenerates significantly.
Tab.~\ref{exp:generated:tab:d10:with:weighted} and \ref{exp:generated:tab:d10:with:unweighted} show that \texttt{SG}$_{cem}$ and \texttt{Ad}$_{cem}$ still work well for $c_w\leq 1$ and, in contrast to data without noise, also for separated instances ($c_w=2$).
\texttt{KG} and \texttt{G}$_{km}$ are now among the methods with the lowest average rank.
This is not a surprise since our noise contains outliers. 
\begin{table*}[htp]
  \centering
   \caption{Average ranks ($\pm$ std.dev.) for generated data with $K=20$, $|X|=1\,000$, dimension $D=10$, \textbf{different weights}, and \textbf{10\% noise}.}
   \label{exp:generated:tab:d10:with:weighted}
    \begin{tabular}{l|ll|ll|ll}
          &  \multicolumn{2}{c|}{\bf separation $c_{\theta}=0.5$}     &  \multicolumn{2}{c|}{\bf separation $c_{\theta}=1$}     &  \multicolumn{2}{c}{\bf separation $c_{\theta}=2$}  \\
          & \multicolumn{1}{c}{initial} & \multicolumn{1}{c|}{final}  & \multicolumn{1}{c}{initial} & \multicolumn{1}{c|}{final} & \multicolumn{1}{c}{initial} & \multicolumn{1}{c}{final} \\ 
\texttt{SG}$\left(s=\frac{1}{10}\right)$ & 8.41$\pm$0.68 & \blue 3.40$\pm$1.75 & 8.22$\pm$0.64 & 4.44$\pm$2.45 & 8.06$\pm$0.68 & 5.46$\pm$2.42 \\
\texttt{SG}$(s=1)$ & 8.25$\pm$0.98 & \blue 3.46$\pm$2.60 & 8.67$\pm$0.47 & 4.13$\pm$3.02 & 8.74$\pm$0.44 & 5.93$\pm$2.87 \\
 \texttt{KG}$(s=1)$ & 10.00$\pm$0.00 & 9.95$\pm$0.22 & 10.00$\pm$0.00 & 9.72$\pm$0.76 & 10.00$\pm$0.00 & 9.02$\pm$1.49 \\
\texttt{Unif}$_{{km}}$ & 1.98$\pm$0.89 & 8.65$\pm$1.03 & 1.05$\pm$0.25 & 7.89$\pm$1.71 & 1.19$\pm$0.49 & 7.34$\pm$1.87 \\
\texttt{G}$_{{km}}$  & 4.29$\pm$1.29 & 5.31$\pm$1.76 & 4.17$\pm$0.98 & 6.59$\pm$1.43 & 3.97$\pm$0.96 & 7.16$\pm$1.38 \\
\texttt{KM++}$_{{km}}$  & 3.23$\pm$0.98 & 6.45$\pm$1.40 & 2.27$\pm$0.60 & 6.83$\pm$1.62 & 2.17$\pm$0.60 & 6.69$\pm$1.69 \\
\texttt{SG}$\left(s=\frac{1}{10}\right)_{{cem}}$  & 6.06$\pm$0.55 & 4.26$\pm$1.36 & 6.04$\pm$0.20 & 3.90$\pm$0.90 & 6.01$\pm$0.091 & 3.65$\pm$1.27 \\
\texttt{SG}$(s=1)_{{cem}}$  & 6.31$\pm$1.43 & 3.49$\pm$2.64 & 7.08$\pm$0.39 & \blue 3.80$\pm$2.89 & 7.19$\pm$0.42 & 5.00$\pm$3.01 \\
 \texttt{Ad}$(\alpha=1)_{{cem}}$ & 3.64$\pm$1.88 & 4.61$\pm$2.26 & 4.05$\pm$0.90 & \blue 3.57$\pm$2.03 & 3.76$\pm$1.26 & \blue 2.06$\pm$1.39 \\
\texttt{Ad}$\left(\alpha=\frac{1}{2}\right)_{{cem}}$ & 2.83$\pm$2.26 & 5.42$\pm$2.70 & 3.46$\pm$0.89 & 4.12$\pm$2.38 & 3.92$\pm$0.78 & \blue 2.69$\pm$1.45 
    \end{tabular}%
\end{table*}%
\begin{table*}[htp]
  \centering
  \caption{Average ranks ($\pm$ std.dev.) for generated data sets with $K=20$, $|X|=1\,000$, dimension $D=10$, \textbf{equal weights}, and \textbf{10\% noise}.}
  \label{exp:generated:tab:d10:with:unweighted}
    \begin{tabular}{l|cc|cc|cc}
          &  \multicolumn{2}{c|}{\bf separation $c_{\theta}=0.5$}     &  \multicolumn{2}{c|}{\bf separation $c_{\theta}=1$}     &  \multicolumn{2}{c}{\bf separation $c_{\theta}=2$}  \\
          & \multicolumn{1}{c}{initial} & \multicolumn{1}{c|}{final}  & \multicolumn{1}{c}{initial} & \multicolumn{1}{c|}{final} & \multicolumn{1}{c}{initial} & \multicolumn{1}{c}{final} \\ 
\texttt{SG}$\left(s=\frac{1}{10}\right)$ & 8.57$\pm$0.62 & 3.38$\pm$1.92 & 8.18$\pm$0.65 & 4.17$\pm$2.28 & 7.94$\pm$0.77 & 5.02$\pm$2.44 \\
\texttt{SG}$(s=1)$ & 8.05$\pm$1.08 & \blue 3.19$\pm$2.23 & 8.68$\pm$0.47 & \blue 3.77$\pm$2.68 & 8.73$\pm$0.44 & 5.47$\pm$2.96 \\
 \texttt{KG}$(s=1)$ & 10.00$\pm$0.00 & 9.93$\pm$0.35 & 10.00$\pm$0.00 & 9.62$\pm$0.87 & 10.00$\pm$0.00 & 7.85$\pm$2.22 \\
\texttt{Unif}$_{{km}}$ & 1.92$\pm$0.87 & 8.83$\pm$0.77 & 1.02$\pm$0.13 & 8.39$\pm$1.22 & 1.11$\pm$0.31 & 8.18$\pm$1.63 \\
\texttt{G}$_{{km}}$  & 4.47$\pm$0.99 & 5.53$\pm$1.75 & 4.01$\pm$1.01 & 6.74$\pm$1.51 & 3.53$\pm$0.83 & 7.18$\pm$1.28 \\
\texttt{KM++}$_{{km}}$  & 3.20$\pm$1.07 & 6.66$\pm$1.22 & 2.08$\pm$0.31 & 7.04$\pm$1.46 & 1.93$\pm$0.37 & 7.66$\pm$1.56 \\
\texttt{SG}$\left(s=\frac{1}{10}\right)_{{cem}}$  & 6.08$\pm$0.53 & 4.47$\pm$1.31 & 6.03$\pm$0.18 & 3.92$\pm$0.97 & 6.00$\pm$0.00 & 3.64$\pm$1.11 \\
\texttt{SG}$(s=1)_{{cem}}$  & 6.20$\pm$1.43 & \blue 3.14$\pm$2.43 & 7.10$\pm$0.40 & 3.88$\pm$2.75 & 7.33$\pm$0.47 & 5.13$\pm$2.76 \\
 \texttt{Ad}$(\alpha=1)_{{cem}}$ & 3.62$\pm$1.85 & 4.42$\pm$2.35 & 4.20$\pm$0.79 & \blue 3.54$\pm$2.11 & 4.15$\pm$0.82 & \blue 2.27$\pm$1.59 \\
\texttt{Ad}$\left(\alpha=\frac{1}{2}\right)_{{cem}}$ & 2.89$\pm$2.45 & 5.45$\pm$2.47 & 3.69$\pm$0.74 & 3.92$\pm$2.29 & 4.28$\pm$0.66 & \blue 2.60$\pm$1.51  
    \end{tabular}%
\end{table*}%
From the results depicted in Tab.~\ref{exp:generated:tab:d10:with:ecc} one can draw the same conclusion, i.e. \texttt{KG} and \texttt{G}$_{km}$ can not handle noisy data.
For noisy data, our \texttt{Ad}$_{cem}$ methods outperform the others.
\begin{table*}[htbp]
  \centering
  \caption{Average ranks ($\pm$ std.dev.) for generated data sets with $K=20$, $|X|=1\,000$, dimension $D=10$, and \textbf{10\% noise}.
  Only final solutions.}
  \label{exp:generated:tab:d10:with:ecc}
\begin{tabular}{l|ccc|ccc}
 & \multicolumn{3}{c|}{\bf equal weights} & \multicolumn{3}{c}{\bf different weights} \\
      & \bf{spherical} & \bf{elliptical} & \bf{both}   & \bf{spherical} & \bf{elliptical} & \bf{both} \\
\texttt{SG}$\left(s=\frac{1}{10}\right)$ & 5.38$\pm$2.26 & 3.79$\pm$2.20 & 4.19$\pm$2.32 & 5.53$\pm$2.30 & 4.07$\pm$2.29 & 4.43$\pm$2.38 \\
\texttt{SG}$(s=1)$ & 4.98$\pm$2.79 & \blue 3.87$\pm$2.76 & 4.14$\pm$2.81 & 5.53$\pm$3.14 & 4.17$\pm$2.90 & 4.51$\pm$3.02 \\
 \texttt{KG}$(s=1)$ & 8.61$\pm$2.03 & 9.31$\pm$1.49 & 9.13$\pm$1.66 & 9.32$\pm$1.28 & 9.64$\pm$0.95 & 9.56$\pm$1.05 \\
\texttt{Unif}$_{{km}}$ & 8.36$\pm$1.36 & 8.50$\pm$1.26 & 8.47$\pm$1.28 & 7.63$\pm$1.72 & 8.07$\pm$1.63 & 7.96$\pm$1.66 \\
\texttt{G}$_{{km}}$  & 6.70$\pm$1.69 & 6.41$\pm$1.67 & 6.49$\pm$1.68 & 6.42$\pm$1.87 & 6.33$\pm$1.66 & 6.35$\pm$1.71 \\
\texttt{KM++}$_{{km}}$  & 7.09$\pm$1.71 & 7.13$\pm$1.39 & 7.12$\pm$1.48 & 6.46$\pm$1.71 & 6.72$\pm$1.53 & 6.66$\pm$1.58 \\
\texttt{SG}$\left(s=\frac{1}{10}\right)_{{cem}}$  & 4.00$\pm$1.45 & 4.01$\pm$1.09 & 4.01$\pm$1.18 & 4.03$\pm$1.34 & 3.90$\pm$1.17 & \blue 3.94$\pm$1.22 \\
\texttt{SG}$(s=1)_{{cem}}$  & 5.04$\pm$2.79 & \blue 3.72$\pm$2.69 & 4.05$\pm$2.77 & 4.88$\pm$2.98 & \blue 3.84$\pm$2.86 & 4.10$\pm$2.92 \\
 \texttt{Ad}$(\alpha=1)_{{cem}}$ & \blue 2.01$\pm$1.29 & \blue 3.88$\pm$2.27 & \blue 3.41$\pm$2.22 & \blue 2.31$\pm$1.57 & \blue 3.78$\pm$2.25 & \blue 3.41$\pm$2.19 \\
\texttt{Ad}$\left(\alpha=\frac{1}{2}\right)_{{cem}}$ & \blue 2.83$\pm$1.72 & 4.37$\pm$2.50 & \blue 3.99$\pm$2.42 & \blue 2.88$\pm$1.78 & 4.48$\pm$2.57 & 4.08$\pm$2.50
\end{tabular}%
\end{table*}

\paragraph{Low Dimensional or High Sample Size Data.}
We expect that, if the dimension is low or the sample size is large enough, it is generally easier to identify clusters.
Indeed the results differ significantly from our previous results. 
In general, the \texttt{KM++}$_{km}$ and \texttt{Unif}$_{km}$ perform best.
For data sets with $D=3$ and $|X|=1\,000$, Tab.~\ref{exp:generated:tab:d3} shows that the \texttt{KM++}$_{km}$ method works well even in the presence of noise.
However, if we are given noise \emph{and} small separation, the simple \texttt{Unif}$_{km}$ does well.
We also increased the sample size to $|X|=5\,000$ \emph{and} the dimension to $D=10$, expecting that the higher sample size can make up for the higher dimension (results available in \cite{SupplementalMaterial}).
Indeed, for data sets without noise, where clusters can presumably be identified easier, \texttt{KM++}$_{km}$ still suffices.
However, given noise or too small a separation, our \texttt{Ad}$_{cem}$ methods and the simple  \texttt{Unif}$_{km}$ work better.
\begin{table}[ht]
  \centering
  \caption{Average ranks ($\pm$ std.dev.) for generated data ($K=20$, $|X|=1\,000$, $D=3$).}
  \label{exp:generated:tab:d3}
    \begin{tabular}{l|rr|rr}
          & \multicolumn{2}{c|}{\bf without noise} & \multicolumn{2}{c}{\bf noisy} \\
          & \multicolumn{1}{c}{initial} & \multicolumn{1}{c|}{final} & \multicolumn{1}{c}{initial} & \multicolumn{1}{c}{final} \\ 
\texttt{SG}$\left(s=\frac{1}{10}\right)$ & 7.31$\pm$0.63 & 7.94$\pm$1.39 & 8.05$\pm$0.70 & 7.93$\pm$1.31 \\
\texttt{SG}$(s=1)$ & 8.90$\pm$0.39 & 8.56$\pm$1.98 & 8.72$\pm$0.45 & 7.99$\pm$2.19 \\
 \texttt{KG}$(s=1)$ & 9.94$\pm$0.42 & 3.28$\pm$1.98 & 10.00$\pm$0.00 & 8.09$\pm$1.46 \\
\texttt{Unif}$_{{km}}$ & 2.82$\pm$0.58 & 4.63$\pm$1.40 & 3.38$\pm$0.94 & 2.76$\pm$1.38 \\
\texttt{G}$_{{km}}$  & 1.93$\pm$1.04 & \blue 2.80$\pm$1.97 & 4.43$\pm$1.59 & 6.01$\pm$1.59 \\
\texttt{KM++}$_{{km}}$  & {1.51$\pm$0.60} & {\blue 1.99$\pm$1.21} & 2.96$\pm$1.12 & \blue 2.35$\pm$1.57 \\
\texttt{SG}$\left(s=\frac{1}{10}\right)_{{cem}}$  & 6.10$\pm$0.42 & 7.46$\pm$1.14 & 5.75$\pm$0.82 & 6.07$\pm$1.16 \\
\texttt{SG}$(s=1)_{{cem}}$  & 7.65$\pm$0.94 & 8.83$\pm$1.74 & 7.04$\pm$0.86 & 8.16$\pm$2.10 \\
 \texttt{Ad}$(\alpha=1)_{{cem}}$ & 4.35$\pm$0.83 & 4.66$\pm$1.48 & 2.54$\pm$1.43 & 3.02$\pm$1.13 \\
\texttt{Ad}$\left(\alpha=\frac{1}{2}\right)_{{cem}}$ & 4.49$\pm$0.74 & 4.85$\pm$1.51 & 2.12$\pm$1.36 & \blue 2.62$\pm$1.28 \\
    \end{tabular}%
\end{table}%

\subsection{Real World Data Sets}\label{sec:experiments-real-world}
We use four publicly available data sets:  
\emph{Covertype} ($|X|=581\,012$, $D=10$ real-valued features)  \cite{Asuncion07}; 
two \emph{Aloi} data sets ($|X|=110\,250$, $D\in\{27,64\}$) based on color histograms in HSV color space \cite{Kriegel11} from data provided by the {ELKI} project\cite{Kriegel12} and the {Amsterdam Library of Object Images}\cite{Geusebroek05}; 
\emph{Cities} ($|X|=135\,082$, $D=2$) is a projection of the coordinates of cities with a population of at least 1000 \cite{Geonames}; 
\emph{Spambase} ($|X|=4601$, $D=10$ real-valued features)\cite{Asuncion07}.

The results are depicted in Figure~\ref{fig:real_world}: 
\begin{figure}
\subfloat[\emph{Aloi} ($D=27$, $K=10$)]{
\includegraphics[width=.3\textwidth]{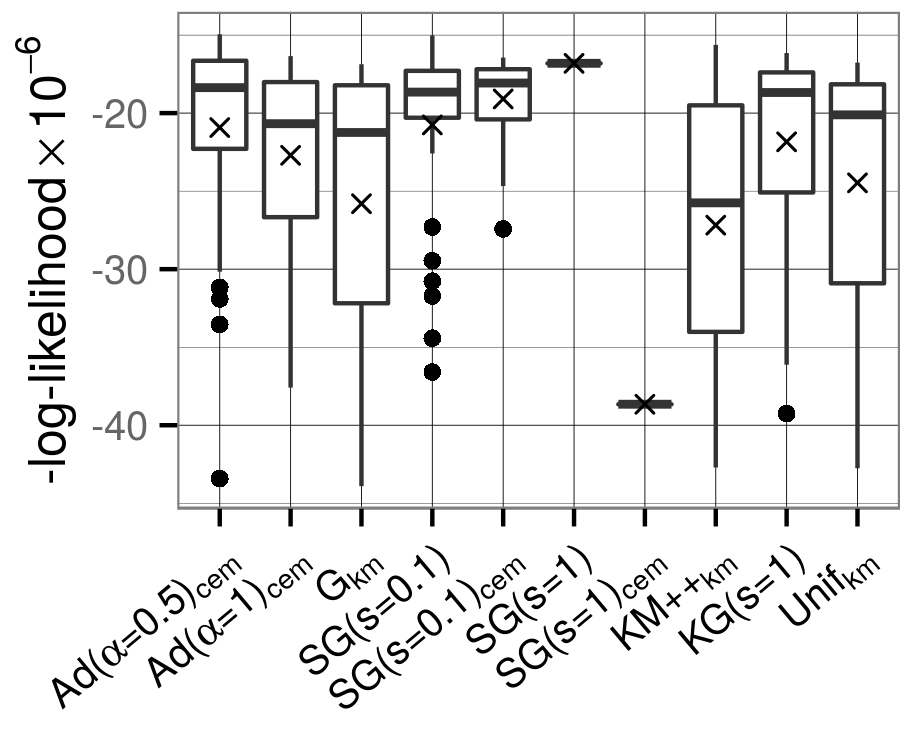}
}\ \ 
\subfloat[\emph{Aloi} ($D=64$, $K=10$, normalized features)]{ 
\includegraphics[width=.3\textwidth]{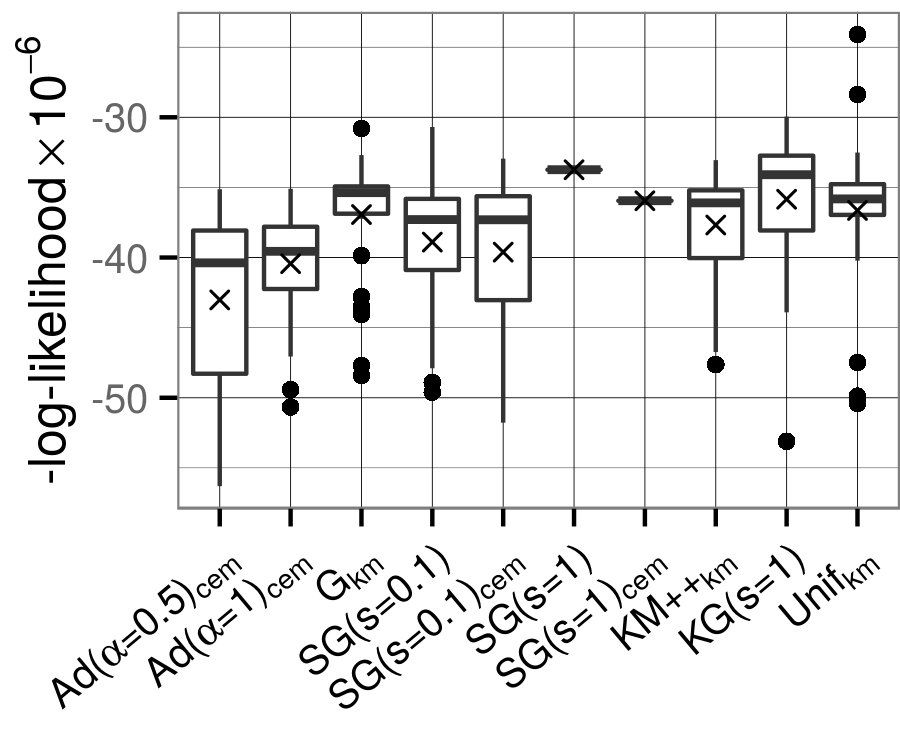}} \ \ 
\subfloat[\emph{Covertype} ($K=10$)]{
\includegraphics[width=.3\textwidth]{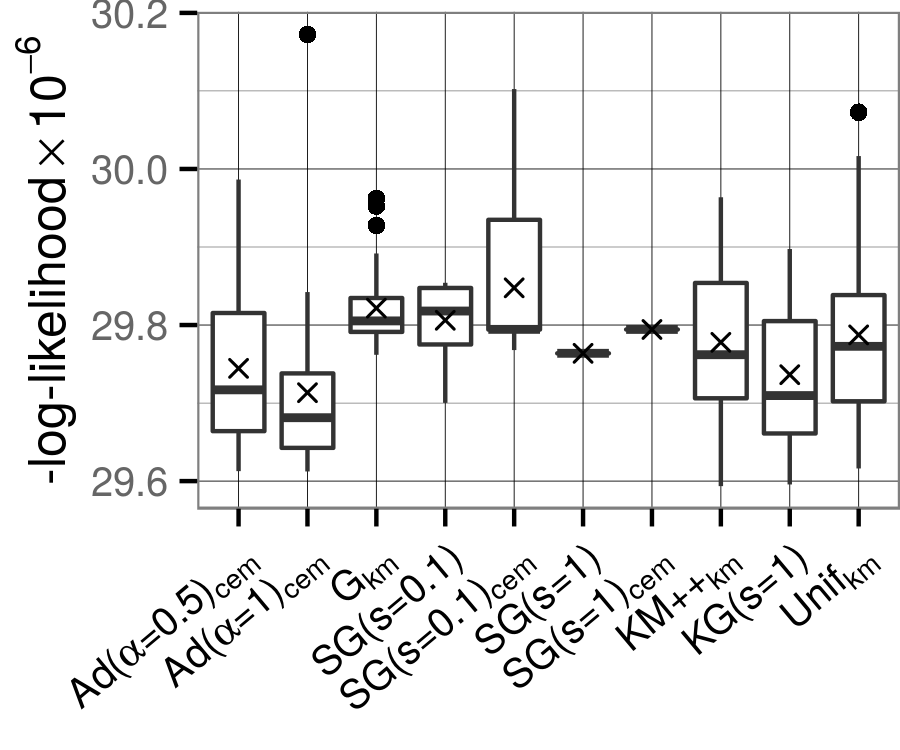}
}

\subfloat[\emph{Spambase} ($K=3$)]{
\includegraphics[width=.3\textwidth]{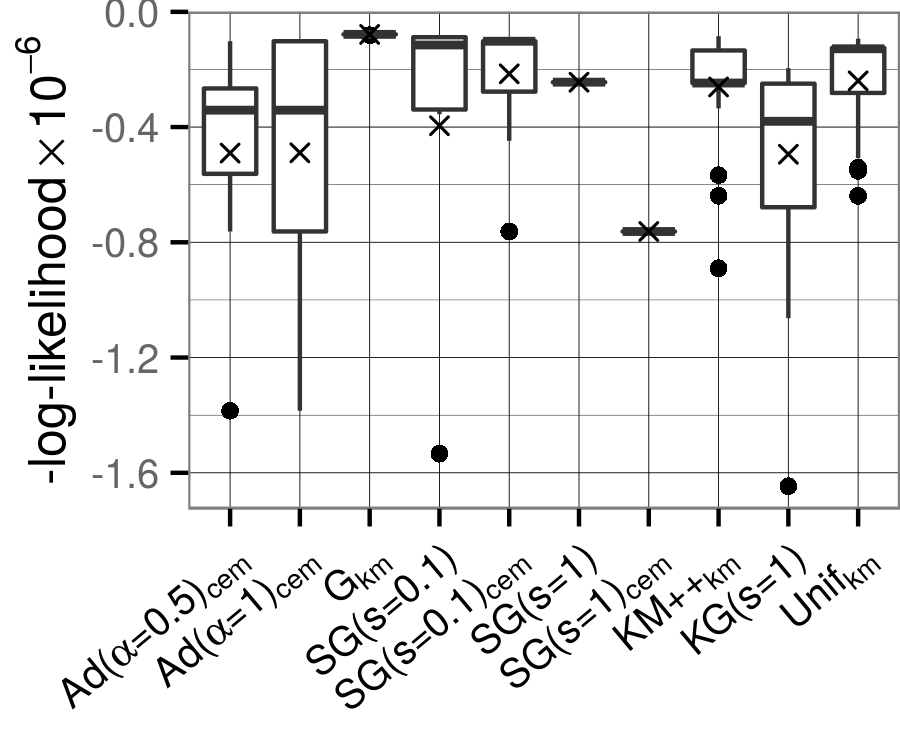}
} \ \  
\subfloat[\emph{Spambase} ($K=10$)]{
\includegraphics[width=.3\textwidth]{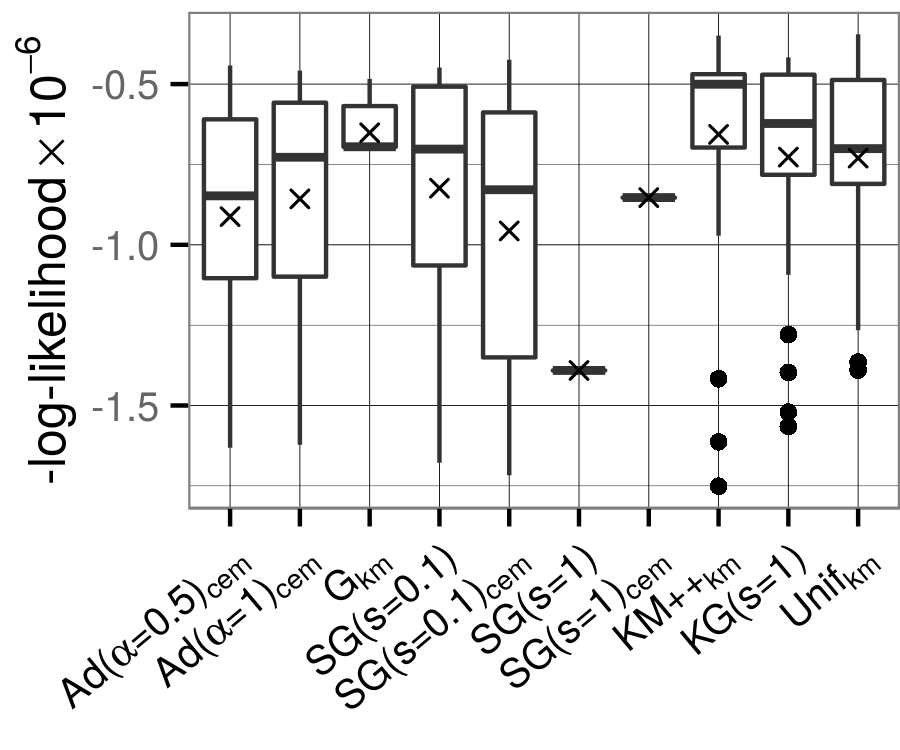}
}\ \ 
\subfloat[\emph{Cities} ($K=10$)]{
\includegraphics[width=.3\textwidth]{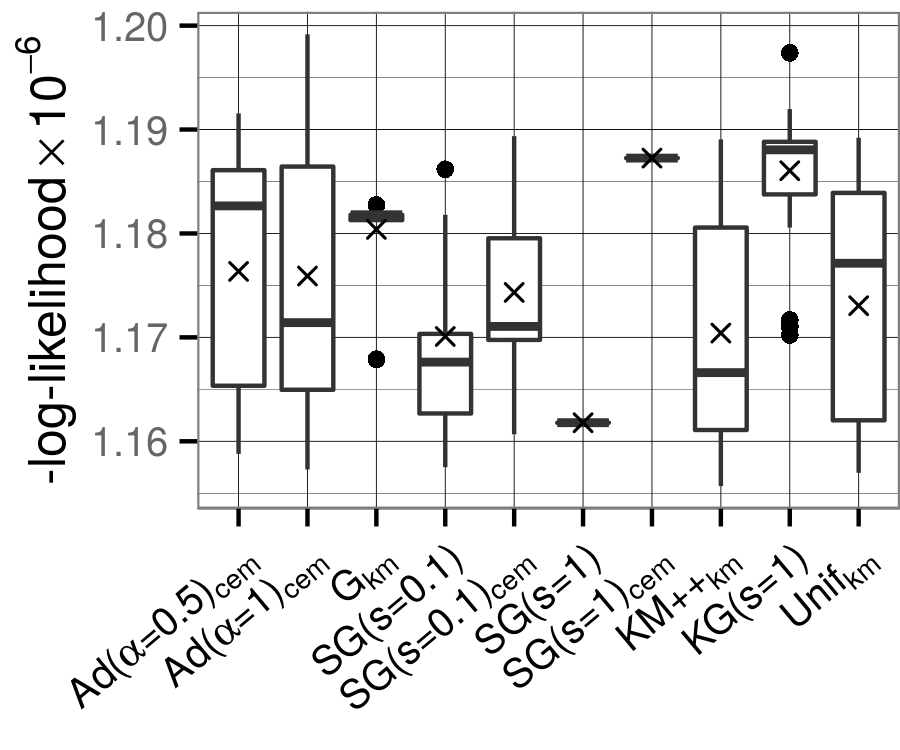}
}
\caption{Results for the real world data sets depicted as boxplots (final solutions only). 
}
\label{fig:real_world}
\end{figure}
For \emph{Aloi} ($D=27$) and \emph{Spambase} ($K=3$), \texttt{SG}$_{cem}(s=1)$ is considerably better than the other methods. 
For  \emph{Cities} and \emph{Spambase} ($K=10$), \texttt{SG}$(s=1)$ does better (without running the CEM). 
For  \emph{Aloi} ($D=64$) and the \emph{Covertype}, \texttt{Ad}$_{cem}(\alpha=1)$ works better than the others.

\subsection{Time Measurement}
The run times of the compared methods match our expectation (cf. Tab.~\ref{tab:time-measurement}).
\begin{table}[htb]
  \centering
  \caption{Average run times (in seconds) over 12 data sets with $|X| = 10^3$ and $D=10$ and different runs per data set using an Intel Core i7-3770 CPU (3.40 GHz, 8GB RAM). }
   \begin{tabular}{rl|rl|rl|rl}
    \texttt{SG}$(s=1)$ & 0.314 
    & \texttt{SG}$(s=0.1)$ & 0.307
    & \texttt{Ad}$_{cem}(\alpha=1)$ & 0.226 
    & \texttt{Unif}$_{km}$ & 0.206 \\
      \texttt{SG}$_{cem}(s=1)$ & 0.225 
     & \texttt{SG}$_{cem}(s=0.1)$ & 0.197
     &\texttt{Ad}$_{cem}(\alpha=0.5)$ & 0.253 
       & \texttt{HAC}$(s=1)$ & 1.588 \\
      \texttt{G}$_{km}$ & 0.205  
     & \texttt{KG}$(s=1)$ & 0.315\\
    \end{tabular}%
  \label{tab:time-measurement}%
\end{table}%
First, (intermediate) steps of the CEM algorithm are faster than (more) steps of the EM algorithm.  
Second, sampling and running methods on a random subset of the data should in general reduce the run time.

\section{Conclusion and Future Work}
If you need a fast and simple method, then we suggest to use one of the following methods:
Given a data set with a large number of points per cluster or low dimension, the $K$-means++ initialization followed by \texttt{Means2GMM} and the $K$-means algorithm should do well.
Otherwise, we recommend our new methods \texttt{Ad} and \texttt{SG} followed by the spherical CEM algorithm, especially if your data is presumably noisy.
Last but not least, whatever you prefer, we suggest trying intermediate steps of the spherical CEM or $K$-means algorithm.

For the $K$-means++ algorithm and the Gonzalez algorithm there are provable guarantees.
We hope that our results are a good starting point for a theoretical analysis that will transfer these results to the MLE problem for GMMs. 

\bibliographystyle{splncs03}
\bibliography{references}

\begin{thebibliography}{10}
\providecommand{\url}[1]{\texttt{#1}}
\providecommand{\urlprefix}{URL }

\bibitem{Geonames}
{GeoNames geographical database}, \url{\url{http://www.geonames.org/}}

\bibitem{Kriegel12}
Achtert, Goldhofer, Kriegel, Schubert, Zimek: {Evaluation of Clusterings --
  Metrics and Visual Support}.
  \url{http://elki.dbs.ifi.lmu.de/wiki/DataSets/MultiView}

\bibitem{ArthurV07}
Arthur, Vassilvitskii: {k-means++: The Advantages of Careful Seeding}. In: SODA
  2007

\bibitem{Asuncion07}
Asuncion: {UCI} machine learning repository (2007),
  \url{http://www.ics.uci.edu/~mlearn/MLRepository.html}

\bibitem{Celeux15}
Baudry., Celeux: {EM for mixtures}. Statistics and Computing  25(4) (2015)

\bibitem{Biernacki04}
Biernacki: {Initializing EM using the properties of its trajectories in
  Gaussian mixtures.} Statistics and Computing  14(3) (2004)

\bibitem{Biernacki03}
Biernacki, Celeux, Govaert: {Choosing starting values for the EM algorithm for
  getting the highest likelihood in multivariate Gaussian mixture models}.
  Comput. Stat. Data Anal.  41(3-4) (2003)

\bibitem{Bishop06}
Bishop: {Pattern Recognition and Machine Learning (Information Science and
  Statistics)}. Springer-Verlag New York, Inc., Secaucus, NJ, USA (2006)

\bibitem{SupplementalMaterial}
Bujna, Kuntze: {Supplemental Material}.
  \url{http://www-old\allowbreak.\allowbreak cs.\allowbreak upb.\allowbreak
  de/\allowbreak fachgebiete/\allowbreak ag-bloemer/\allowbreak
  forschung/\allowbreak clusteranalyse/\allowbreak
  adaptive\_seeding\_for\_gmms.html}

\bibitem{Celeux92}
Celeux, Govaert: {A Classification EM Algorithm for Clustering and Two
  Stochastic Versions}. Comput. Stat. Data Anal.  14(3) (1992)

\bibitem{Dasgupta00}
Dasgupta: Experiments with random projection. In: {UAI} '00

\bibitem{Dasgupta99}
Dasgupta: {Learning Mixtures of Gaussians}. In: FOCS 1999

\bibitem{Dempster77}
Dempster, Laird, Rubin: {Maximum likelihood from incomplete data via the EM
  algorithm}. {Journal of the Royal Statistical Society, Series B: Statistical
  Methodology}  39(1) (1977)

\bibitem{FGK}
{F\"arber, G\"unnemann, Kriegel, Kr\"oger, M\"uller, Schubert, Seidl, and
  Zimek}: {On Using Class-Labels in Evaluation of Clusterings}. In: MultiClust
  2010

\bibitem{Fayyad98}
Fayyad, Reina, Bradley: {Initialization of Iterative Refinement Clustering
  Algorithms.} In: KDD 1998

\bibitem{Geusebroek05}
{Geusebroek, Burghouts, and Smeulders}: {The Amsterdam Library of Object
  Images}. International Journal of Computer Vision  6(1)

\bibitem{Gonzalez85}
Gonzalez: Clustering to minimize the maximum intercluster distance. Theoretical
  Computer Science  38 (1985)

\bibitem{Kriegel11}
Kriegel, Schubert, Zimek: {Evaluation of Multiple Clustering Solutions}. In:
  MultiClust 2010

\bibitem{Krueger10}
Kr{\"u}ger, Leutnant, Haeb-Umbach, Ackermann, Bl\"omer, J.: {On the
  initialization of dynamic models for speech features}. Sprachkommunikation
  2010

\bibitem{Kwedlo2015}
Kwedlo: {A new random approach for initialization of the multiple restart EM
  algorithm for Gaussian model-based clustering}. Pattern Analysis and
  Applications  18(4),  757--770 (2015)

\bibitem{Maitra09}
Maitra: {Initializing Partition-Optimization Algorithms}. IEEE/ACM Transactions
  on Computational Biology and Bioinformatics  6(1) (2009)

\bibitem{Maitra10}
Maitra, Melnykov: {Simulating data to study performance of finite mixture
  modeling and clustering algorithms}. Journal of Computational and Graphical
  Statistics  19(2) (2010)

\bibitem{Mclachlan08}
McLachlan, Krishnan: {The EM Algorithm and Extensions (Wiley Series in
  Probability and Statistics)}. Wiley-Interscience, 2 edn. (2008)

\bibitem{Meila98}
Meil{\u{a}}, Heckerman: {An Experimental Comparison of Several Clustering and
  Initialization Methods}. In: UAI 1998. Morgan Kaufmann, Inc., San Francisco,
  CA

\bibitem{Melnykov11}
Melnykov, Melnykov: {Initializing the EM algorithm in Gaussian mixture models
  with an unknown number of components}. Computational Statistics \& Data
  Analysis  (2011)

\bibitem{Thiesson98}
Thiesson: {Accelerated quantification of Bayesian networks with incomplete
  data}. University of Aalborg (1995)

\bibitem{Verbeek03}
Verbeek, Vlassis, Kr{\"o}se: {Efficient greedy learning of Gaussian mixture
  models}. Neural computation  15(2) (2003)

\end{thebibliography}

\end{document}